\documentclass[letterpaper]{article}
\usepackage{booktabs}
\usepackage{aaai}
\usepackage{multicol}
\usepackage{times}
\usepackage{helvet}
\usepackage{courier}
\usepackage{xcolor}

\newtheorem{defn}{Definition}[section]

\usepackage{fullpage,graphicx,psfrag,amsmath,amsfonts,verbatim}

\usepackage{float}
\usepackage[small,bf]{caption}
\usepackage{subcaption}
\usepackage{booktabs}
\begin{document}
%
\title{Incentivizing an Unknown Crowd}

\author{Jing Dong,\textsuperscript{1}
Shuai Li,\textsuperscript{2}
Baoxiang Wang,\textsuperscript{1}\\
\textsuperscript{1}{The Chinese University of Hong Kong, Shenzhen}\\
\textsuperscript{2}{Shanghai Jiao Tong University}\\
jingdong@link.cuhk.edu.cn,
shuaili8@sjtu.edu.cn, 
bxiangwang@cuhk.edu.cn}

\maketitle
\begin{abstract}
Motivated by the common strategic activities in crowdsourcing labeling, we study the problem of sequential eliciting information without verification (EIWV) for workers with a heterogeneous and unknown crowd. We propose a reinforcement learning-based approach that is effective against a wide range of settings including potential irrationality and collusion among workers. With the aid of a costly oracle and the inference method, our approach dynamically decides the oracle calls and gains robustness even under the presence of frequent collusion activities. Extensive experiments show the advantage of our approach. Our results also present the first comprehensive experiments of EIWV on large-scale real datasets and the first thorough study of the effects of environmental variables. 
\end{abstract}

\section{Introduction}
The increase in the model complexity of machine learning methods poses new challenges to data acquisition, as obtaining high-quality labeled data is often expensive and time-consuming. Crowdsourcing is proposed to use crowd's wisdom to obtain a large volume of labeled data \cite{howe2006rise,simpson2015language,simpson2015language,difallah2015dynamics}.
The crowdsourcing process can be depicted by a Stackelberg game consisting of two parties, the crowdsourcing workers and the platform \cite{cruz1975survey}. The game proceeds in an iterative manner, where at each round the platform first posts a price for labeling and the workers then complete the labeling task in a way best to their interest. 
To obtain high-quality data, this group of workers has to be properly incentivized. 
With the presence of ground truth labels, it amounts to the incentive mechanism design and has been extensively investigated in the literature \cite{pmlr-v37-shaha15,shah2015double,pmlr-v48-shaha16}. Yet in most scenarios, ground truth labels are not readily available. This leads to the problem of eliciting information without verification (EIWV), which is first formulated by \cite{witkowski2013learning} and later discussed from the perspectives across multiple domains, ranging from game theory, data inference, to machine learning 
\cite{dasgupta2013crowdsourced,liu2016learning,liu2017sequential,hu2018inference}. 

In this paper, we focus on the modeling and empirical performance of EIWV and propose a reinforcement learning approach to design an adaptive and effective incentive mechanism. 
Our study of EIWV is with dynamic and online interactions to corroborate real applications.
We study the sequential EIWV problem, in which the crowd of workers strategically label assigned tasks (at some effort cost and payment) to maximize their overall utility.
Stringent assumptions on the crowd of workers and the crowdsourcing tasks that are typical for mathematical tractability are removed. 
The adaptiveness of workers and arbitrarity of tasks then require the crowdsourcing platform to promote honest behavior via payment rule design. This arises the difficulty of inferring ground-truth labels among strategic responses.

Despite the community's efforts on EIWV, relevant literature focus on theoretical guarantees of their approaches while imposing strong assumptions. Their approaches are thus under certain limitations in real applications. 
For example, assumptions like fixed and known cost and full rationality of workers are hard to satisfy in practice \cite{10.1145/2229012.2229085,liu2016learning,hu2018inference}. Without extensive empirical evaluation on real dataset, the robustness of the approach to violations of the assumptions are not verified. 
Moreover, existing approaches often focus on particular game-theoretic formulations, which are hard to be generalized to other settings and are difficult to scale. For mathematical tractability, previous theoretical analyses are often restricted to binary labeled tasks and binary effort labels, where both are relaxed in our setting. Meanwhile, workers are assumed to possess the ability to exactly optimize their utility, which does not hold in real settings \cite{witkowski2013dwelling,10.1145/2940716.2940790,liu2017sequential}. 
The above limitations indicate an immediate need in designing a practical mechanism that is robust in a variety of regimes regardless of worker rationality and is flexible to adapt to different settings with easiness to scale.

Incentivizing an unknown crowd of workers is hard due to not only the potential irrationality of workers but also potential collusion among them. In the most extreme case where all workers collude together, it is infeasible to detect the collusion in a provably sound way. We show that by leveraging only a costly oracle, as a practical implementation of the ``golden label'' studied in the literature \cite{10.1145/3396863,8693123}, the effect of collusion can be mitigated. This is applicable even when the cost of the oracle query is very high, which makes the approach practical when golden labels are scarce and when third-party experts are expensive. 
With the presence of collusion, the design of the incentive mechanism is further complicated by the additional need for a dynamic strategy of oracle queries.

Our approach formulates the sequential EIWV problem into a decision problem as a Markov decision process and proposes a reinforcement learning-based method to design an effective incentive mechanism. To tackle the problem of inferring true labels among strategic responses, our method adapts available inference methods with the aid of a costly ground truth oracle. We summarize our contributions as follows.
\begin{enumerate}
    \item We propose an incentive mechanism design framework for EIWV with an unknown but potentially diverse and strategic crowd of workers. The framework is aware that workers may not be fully rational and have arbitrary effort cost distribution. With the aid of a costly oracle, our approach achieves high utility and is robust to worker collusion. 
    \item Our proposed framework is effective in a wide range of settings with different types of tasks.
    \item We provide the first comprehensive empirical analysis of EIWV on real-world datasets and the first study of the practical implications of environmental variables, such as task assignments, cost distributions, strategic behavior, etc.
\end{enumerate}

\section{Related Work}

The challenge of incentivizing workers to invest high efforts while promoting crowd label accuracy without the ground truth, formally known as eliciting information without verification (EIWV), is first introduced by \cite{witkowski2013learning} and \cite{dasgupta2013crowdsourced}.
Under binary effort cost level and binary labeled tasks, both work proposed one-shot EIWV mechanisms such that exerting high effort and reporting truthfully about workers' observations is a Nash equilibrium. 
While the results are appealing, their approach has a few limitations. 
The effort cost is assumed to be homogeneous across the crowed while being fixed and known, which is impractical in many real applications. The workers are also assumed to be fully rational and the tasks labels are also assumed to be binary.
Beyond the one-shot setting, \cite{liu2016learning} studied a sequential version of EIWV and proposed a sequential mechanism to learn the optimal payment rule. 
In the study, it requires the workers to report their effort costs along with their answers, which is not feasible in most real applications. 
Moreover, their mechanism is not resistant to even simple collusion among workers. 
The framework is later extended to the case without the knowledge of worker's effort cost in \cite{liu2017sequential}. They proposed a sequential peer prediction mechanism, which is collusion resistant under binary labeled tasks and binary but heterogeneous effort cost levels. 
For this peer prediction method to be effective, each worker must label the same number of tasks at each round. 
The results are later generalized to multi-class labels by \cite{shnayder2016informed} and \cite{10.1145/3296670}.

\citeauthor{han2021truthful} (\citeyear{han2021truthful}) and \citeauthor{10.1145/3442381.3449840} (\citeyear{10.1145/3442381.3449840}) considered settings where workers are from a hybrid crowd, with potentially honest workers, rational workers, and adversarial workers. 
The former focuses only on the binary labeled tasks, while the latter employs a robust learning approach to elicit information beyond binary labeled tasks. 
This setting remains to be different from ours as we consider collusion among workers rather than potentially adversarial workers (the workers are to their interests instead of to adversarially jeopardize the algorithm). In fact, under adversarial setting a incentive mechanism's effect will be limited. 
In contrast, if a subset of workers decide to collude, there may exist an effective payment to incentive the workers by stopping the collusion.

The first to employ reinforcement learning on EIWV is  \cite{hu2018inference}. 
With Gaussian process Q-learning and Gibbs sampling, their method is both incentive-compatible and individual-rational for workers to exert high effort and report truthfully under binary answer space and binary effort level. 
This implies that it is in the best interest of rational workers to exert high effort and report truthfully for better rewards. 
Besides the limitation incurred by the simple answer space and effort level assumptions, their mechanism is empirically analyzed through only one real dataset under the fulfilled incentive-compatible requirement. It is worth pointing out that \citeauthor{hu2018inference} \citeyear{hu2018inference} also discusses a trick to build up signal for collusion, which works if all workers collude for an extended period of time.

\section{Preliminaries}

We model a crowdsourcing data acquisition task as a strategic game between workers and the crowdsourcing platform. Through a finite time horizon $T$, $N$ workers will label $m_t$ tasks at each time step $t$. 
The game proceeds as workers observe a posted payment for each task at each time step, then they modify their labeling and effort exertion strategy to maximize their utility. While the workers seek to maximize their utility, the crowdsourcing platform needs to balance between maximizing labels' quality and minimizing payment spent.

We first give the definition of workers' and the platform's utility, which is general enough to be straightforwardly adapted to a variety of settings.
Let worker $i$'s effort level at time $t$ be $e_{i,t}$, which is at least its base effort cost level. We do not impose assumptions on the workers' background and thus they may possess diverse skills. This results in various base effort costs among workers, representing various capability of completing the tasks. Note that the true distribution of this base effort cost level remains unknown to the platform. Let total payment to worker $i$ at time $t$ be $P_{i,t}$ and the number of tasks completed by worker $i$ at $t$ be $m_{i,t}$. The utility ${U}_{w,i}$ of a worker is defined as 
\begin{align}
\label{utility_w}
    {U}_{w,i} = \mathbb{E} \left[\sum^T_{t=0} \left(P_{i,t} - e_{i,t} \cdot m_{i,t}\right)\right] \,.
\end{align} 
On the platform side, it needs to balance between accurate results and the payment spent through a variable parameter $\eta$ as the payment weight. Let the total number of tasks completed at time $t$ be $m_t$, which is the sum of all tasks completed by the workers ($m_t = \sum_{i \in N}m_{i,t}$). Denote the accuracy
of a label $m_t$ as $A_t$. Then the platform's utility, denoted by ${U}_p$, is defined as
\begin{align}
\label{utilityp}
    {U}_p = \mathbb{E} \left[\sum^T_{t=0} \left( A_t - \eta \cdot \sum^N_{i=1} P_{i,t} \right)\right]\,.
\end{align}
Without the ground truth information about the labels, the platform only has limited access to the value of $A_t$ through an inference mechanism (e.g. an EM algorithm \citeauthor{10.14778/3055540.3055547} \citeyear{10.14778/3055540.3055547}) that provides an estimation for label accuracy. Let $\hat{A}_t$ be the accuracy that the platform observes through the inference mechanism. The platform then aims to maximize $\hat{U}_p = \mathbb{E} \left[\sum^T_{t=0} \hat{A}_t - \eta \cdot P_t \right]$.

With the utility definitions, 
we consider the interactions between the two parties by not only considering independent working strategies \cite{dasgupta2013crowdsourced,hu2018inference}, but also collusion among workers. When collusion takes place, a subset of workers works together to modify their strategy according to the group interest. Formally, we define collusion for rational workers as follows.
\begin{defn}[Collusion for rational workers]
    Given a population $[N]$ of rational workers and a subset of workers $G \subseteq [N]$. For workers in $[G]$ and some assigned task, let $r_i$ be the honest response for worker $i\in[N]$ and $r=\{r_i, i\in G\}$ the honest response of the group of workers. If $G$ colludes, workers in $G$ will not submit $r_i$ if there exists an $r'$ such that $U_{w,i}(r) < U_{w,i}(r')$, $\forall i \in G$.
\end{defn}

When workers are fully rational, existing works on peer prediction has effective ways to prevent collusion \cite{liu2017sequential}. However, in sequential EIWV, it is hard in general for workers to exactly optimize towards the expected utility. Thus collusion becomes a much likely event and its occurrence brings catastrophic damage to the utility of the platform. 
Intuitively some payment rules can stop the collusion, for example, one that favors honest and non-collusion participants. Yet the crowdsourcing platform needs to identify collusion and properly incentivize workers with very limited information. It is immediate that without ground truth labels, collusion is impossible to detect. We thus propose an oracle-aided approach to this problem, where the definition of an oracle is formally given as follows.
\begin{defn}[$\alpha$-cost oracle]
    For a given payment $P_t$, and when an $\alpha$-cost oracle is called, the crowdsourcing platform receives a signal $S_t$ which encodes the true effort level of workers at the cost of paying $P_t ( 1 + \alpha)$ instead. 
\end{defn}
Under our setting, the signal $S_t$ may be the true accuracy of the worker's labeling service, which encodes information about the worker's actual effort level. The oracle may be implemented in several ways in real settings. In the offline case it corresponds to a partially labeled dataset, where the fraction of tasks with ground truth labels can be utilized as the oracle. In an online case the $\alpha$-cost oracle can be intuitively understood as the option to resort to a trusted expert that is $\alpha$ more expensive than a normal worker. This costly oracle is realistic in general and is available in many real applications \cite{oleson2011programmatic,shah2015double,8693123,checco2019quality}.

\begin{figure}[H]
    \centering
    \includegraphics[width=0.5\textwidth]{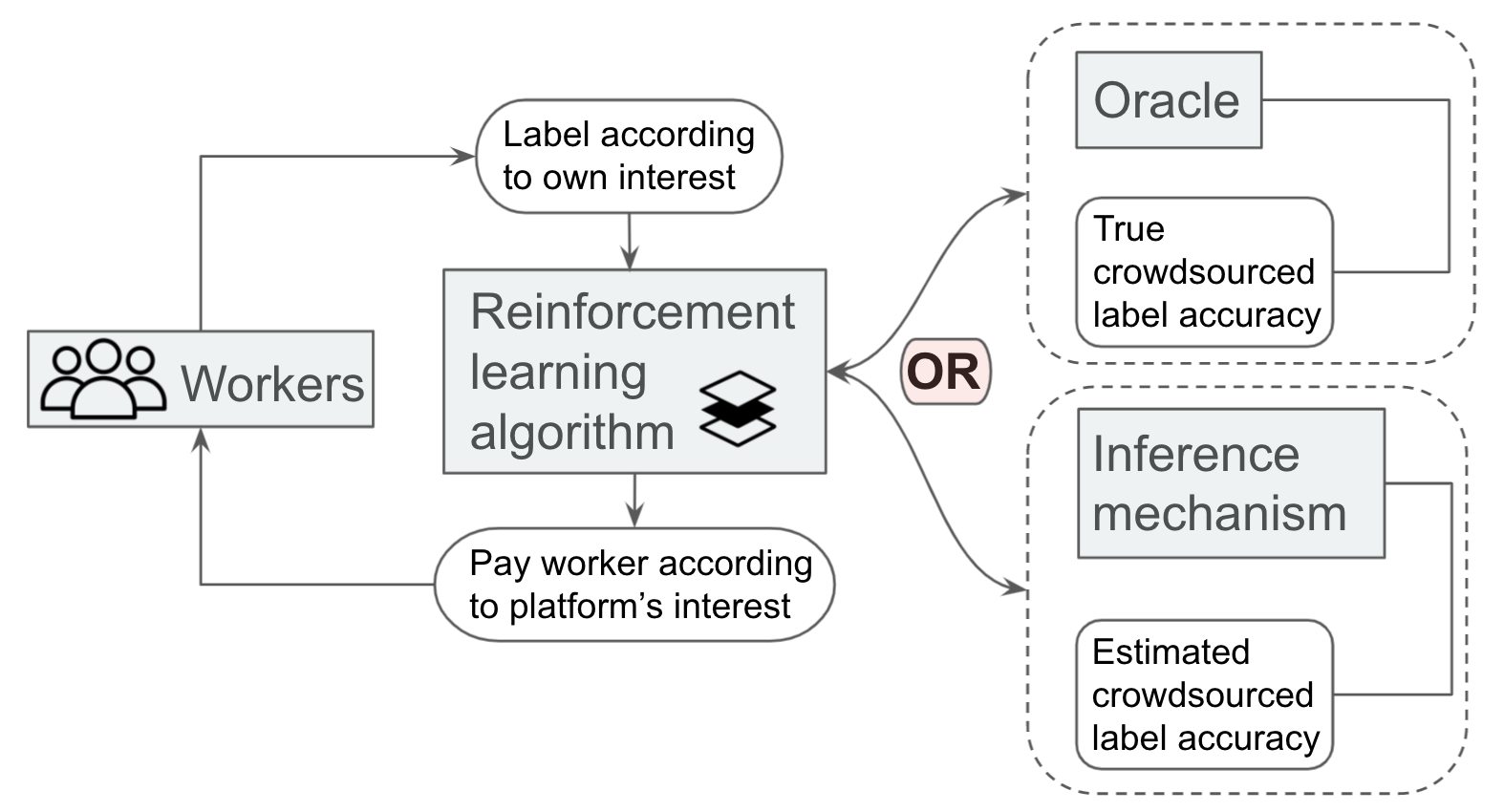}
    \caption{Flowchart of our framework}
    \label{fig:flow}
\end{figure}

With the described setting, we use reinforcement learning to derive an incentive payment policy. To cope with potentially adversarial or irrational behavior from workers, we let reinforcement learning to learn the choice of calling a costly but accurate oracle, as a substitute for low-cost but potentially inaccurate inference mechanism (e.g. an EM algorithm). The interactions between workers and the underlying reinforcement learning can be captured by Figure \ref{fig:flow}.

\section{Reinforcement Learning for EIWV}

We first translate the setting of sequential EIWV into a Markov decision process (MDP) denoted by the tuple $\mathcal{M} = (\mathcal{S}, \mathcal{A}, \mathcal{R}, \mathcal{P}, \gamma)$, where $\mathcal{S}$ is the state space, $\mathcal{A}$ is the action space, and $\gamma$ is the discount factor. For an $(s, a, s')$ tuple where $s, s' \in \mathcal{S}, a \in \mathcal{A}$, $\mathcal{R}(s,a)$ is the reward function and $\mathcal{P}(s, a, s')$ describes the transition probability of the MDP. For our problem, let the accuracy of workers be the observation state $s$ and let the payment to the workers be the action $a$. Specifically, at time $t$, the state is expressed as $\hat{s}_t = (\hat{s}_{1,t}, \dots, \hat{s}_{N,t})$, which is a vector representation for estimated accuracy of worker $1, \dots, N$ based on the inference method of the platform's choice. The action is also a vector representation of the payments for the next batch of tasks to worker $1, \dots, N$, represented as $a_t = (a_{1,t}, \dots, a_{N,t})$. The action space can be either discrete or continuous, where a continuous action space 
allows more flexible mechanism design but poses greater difficulty for RL algorithms. The state transition is governed by the workers' adaptive labelling strategies and remains unknown to the RL algorithm. We define the reward function to be the running average of the platform utility \eqref{utilityp} and the algorithm's goal is to maximize the cumulative reward. 

The objective of the RL algorithm is to learn a policy $\pi: \mathcal{S} \to \mathcal{A}$ that maximizes the cumulative rewards over a finite horizon $T$. The RL algorithm is provided with an $\alpha$-cost oracle as an optional alternative to the inference method. We augment this choice of calling oracle into the MDP action space with a simple trick. Specifically, when the RL algorithm decides to give the lowest possible payment to a larger portion than some threshold of the population, we regard that the RL algorithm is uncertain of its payment. 
An oracle will be called then at an extra cost and the overall effect of using the oracle is reflected on the received reward. 

We adapt a popular RL algorithm, A2C, as our learning algorithm \cite{mnih2016asynchronous}. The A2C algorithm consists of an actor and a critic, where the actor aims to optimize the policy incorporating the information from the critic, and the critic evaluates the policy. When collusion is present and without an always available oracle, directly employing A2C with other popular inference methods (e.g. an EM algorithm) yields suboptimal performance caused by unstable training (see experiments). When the oracle is available as an option, we refer to this as partial oracle support. Under the existence of a partial oracle, we introduce an importance sampling-aided buffer to stabilize the training and further improve the performance.

\subsection{Importance sampled experience replay}

Collusion or irrational behavior may only appear for a limited fraction of timesteps but can cause catastrophic damage. Thus we desire the RL approach for EIWV to be robust to collusion and irrational workers with only limited experiences that are relevant to these situations. Without additional measure, due to the rarity of collusion and irrational activities, the RL agent has limited opportunity to learn its response during training. Intuitively, samples where collusion and irrational behavior happen should be emphasized and learned with high priority. 

For existing popular off-policy RL algorithms, such as off-policy actor-critic \cite{konda2000actor,haarnoja2018soft}, experience replay plays an importance role to help stabilize training and to improve sample complexity. Prioritized experience replay (PER), proposed by \cite{schaul2015prioritized}, leverages the advantage of prioritizing sampling transitions based on their TD error to improve RL algorithm's performance. 
Borrowing this idea, We re-weight the transitions by the likelihood of a collusion and irrational activities occur in and around it. 

It is important to observe that when collusion and irrational activities occur, it usually comes with a sudden change in the observations received by the algorithm. Taking collusion as an example, in the case where a sufficiently large proportion of workers participate in the collusion and the inference method was mislead to the drastically different prediction, the RL algorithm may experience sudden changes of both the inferred true label and the inferred accuracy. In the extreme case where all workers coordinate their answers, the RL algorithm will receive a full accuracy for each entry in its observation state. Note that by describing the change being sudden, we are comparing it to observation state changed caused by the payment policy, which is more gradual. 

We thus use this fact to prioritize importance transitions in the experience replay by importance sampling techniques. Denote the diameter in observation state by $\Delta$, which is the maximum possible distance between two arbitrary states, and let $\delta_t$ be the difference between the observation states $s_t$ and $s_{t+1}$. When transitions are sampled from the buffer during training, each transition is sampled with probability $\delta_t / \Delta $.

\section{Empirical Findings}

We now give a description over the implementation details and the empirical results of our proposed framework for EIWV. We conduct our experiments over 4 datasets that contain a variety of tasks, including binary and multi-label tasks, with varying sizes. we defer the discussion of the flexibility of our algorithm under different environment configuration and a comprehensive list of dataset description to the appendix.

\subsection{Framework for EIWV}

To facilitate the empirical study of EIWV, we provide the first flexible framework for the task and will verify the flexibility and the performance in this section. Our framework can easily incorporate various peer predictions and inference methods, as well as different incentive mechanism including learning based ones. We highlight the following flexibility of our framework.
\begin{enumerate}
    \item \textbf{Dataset:} We support both binary and multi-class, discrete and continuous labeled tasks. If the framework is used without oracle, then the dataset needs no ground truth labels. 
    \item \textbf{Crowd:} The framework is flexible with different crowd settings such as the cost distribution among crowd, the rationality, and the crowd labelling strategy. 
    \item \textbf{Inference method and incentive mechanism:} The framework works with any inference and peer prediction methods and various incentive mechanisms. 
\end{enumerate}

\subsection{Experimental details}


\paragraph{Effort cost distribution} 
Previous works such as \cite{liu2016learning} and \cite{liu2017sequential} require either known worker costs or the costs to be reported by the workers. In contrast, we consider incentivizing workers without the knowledge about their costs and regard such costs as private information of the workers. While previous empirical analysis which focuses on binary effort cost (e.g. either low or high) \cite{hu2018inference}, we model the effort cost level to be in the range of $[h]$ (set as $h=10$ through the experiments). To model the effort cost distribution, our framework takes an upper bound on the effort cost and uniformly randomly initiates an effort cost (without loss of generality, larger than 1) for each worker. 
This distribution remains unknown to the incentive mechanism throughout the interactions, though it remains stationary. This modeling allows the framework to cover a significantly more complicated worker crowd and thus represents a richer class of EIWV problems. 

\paragraph{Task assignment}
A significant line of work \cite{ho2012online,pmlr-v28-ho13,zhao2020preference} studies the optimal task assignment when tasks are heterogeneous. As optimal task assignment is out of the scope of this paper, during training and testing, our framework samples a specified amount of tasks uniformly randomly from all available tasks. 
The task is considered to be assigned to a worker if this worker has labeled the task in the dataset. Note that the number of tasks assigned to each worker can therefore be different. 
We should expect that with a more sophisticated task assignment strategy, the mechanism will perform better than under our framework.

\paragraph{Worker strategy}
Previous models such as \cite{liu2017sequential} and \cite{hu2018inference} assume that rational workers have the ability to exactly maximize their utility function. However, in real settings, rational workers may have limited information to achieve so. Thus we model the workers to be greedy and only optimize a one-step utility function $\mathbb{E} \left[ P_t - e_{i,t} \cdot m_{i,t} \right]$ with an $\epsilon$ accuracy. Specifically, worker will exert effort $\hat{e}_{i,t}$ such that $\left| \mathbb{E} \left[\hat{e}_{i,t} - e^\ast_{i,t} \right] \right| \leq \epsilon$, where $e^\ast_{i,t} = \arg \max \mathbb{E} \left[ P_{i,t} - e_{i,t} \cdot m_{i,t}\right]$.

\begin{figure*}[!htbp]
\begin{minipage}[t]{0.25\textwidth}
  \includegraphics[width=0.95\linewidth]{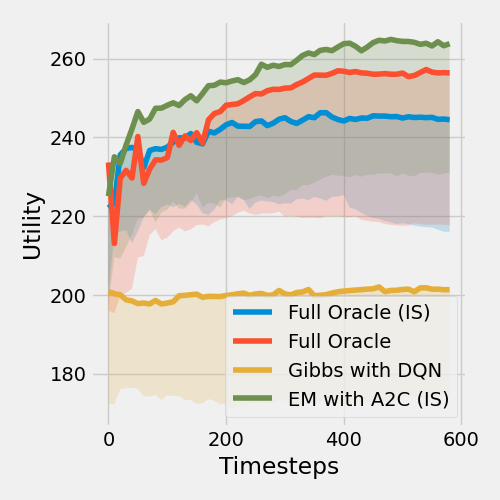}
  \subcaption{}
  \label{fig:compare_base}
\end{minipage}%
\hfill
\begin{minipage}[t]{0.25\textwidth}
  \includegraphics[width=0.95\linewidth]{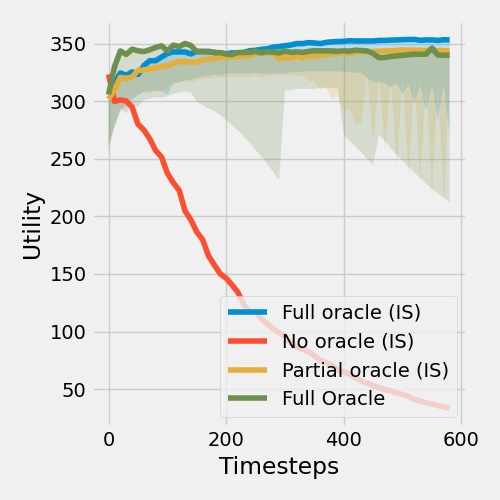}
  \subcaption{}
  \label{fig:amazon_is}
\end{minipage}%
\hfill
\begin{minipage}[t]{0.25\textwidth}
  \includegraphics[width=0.95\linewidth]{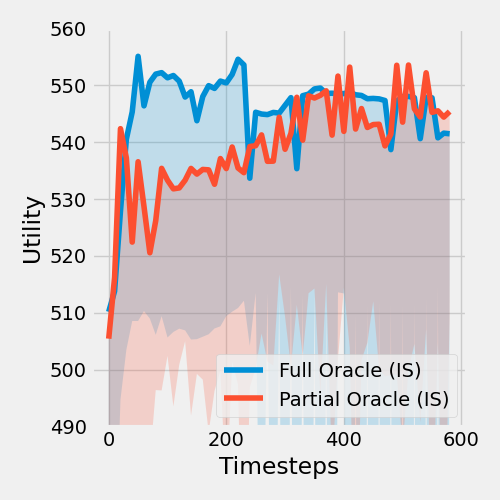}
  \subcaption{}
  \label{fig:binaryAMT}
\end{minipage}%
\hfill
\begin{minipage}[t]{0.25\textwidth}
  \includegraphics[width=0.95\linewidth]{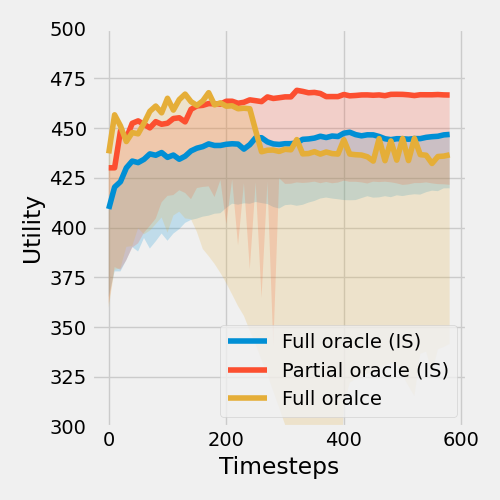}
  \subcaption{}
  \label{fig:amt_multi}
\end{minipage}%
\caption{All figures are results of experiments conducted over 600 timesteps, collusion rate of 50, a maximum payment of 11, maximum effort cost of 10, and 10 tasks per time step. The reward is calculated by the rolling average of the utility in a window of the recent 200 timesteps. The environment configuration remains the same in other figures unless otherwise indicated. Figure \ref{fig:compare_base} compares the performance of our mechanism with and without importance sampled buffer with the Gibbs sampling and Q-learning \cite{hu2018inference} on Bluebirds dataset without collusion. Figure \ref{fig:amazon_is} compares the performance with/without oracle on crowdsourced amazon dataset with a collusion rate of 150 and 100 tasks per timestep. Figure \ref{fig:binaryAMT} compares the performance of A2C with and without an oracle with the Sentiment Popularity - AMT dataset. Figure \ref{fig:amt_multi} compares the performance of mechanism with A2C with the multi-label Weather Sentiment - AMT dataset. }
\label{fig:all_dataset}
\end{figure*}

\paragraph{Other experimental settings}
The performance of the mechanism is evaluated by the cumulative reward, where the reward is defined as a running average of utility over the timesteps. We employ the classic EM algorithm implemented by \cite{10.14778/3055540.3055547} for our framework, but the framework can be easily adapted to other inference methods. The collusion activities is described by the collusion rate of $m$. The workers conduct a collusion every $m$ timesteps and can only be stopped by giving every participant the lowest possible payment. For each figure we present, we repeated with 5 different random seeds for better reproducibility.

\subsection{Results}

We first evaluate our importance sampled experience replay with A2C against the previous approach, its naive counterpart (without importance sampling), A2C with full oracle, and A2C without full oracles with different datasets. 
Due to the setting limitations of previous works, in which no collusion and multi-labeled tasks are considered, this result is obtained through interacting with a fully rational crowd with no possible collusion on the Bluebirds dataset. The previous approach is also only feasible for a discrete action space and a uniform payment rule for all workers (that is, every worker will receive the same amount of payment at each timestep), we thus discretize the continuous action space used for our approach for comparison. Among previous approaches, the closest to our setting is the Q-learning approach with Gibbs sampling (Gibbs with DQN) from \cite{hu2018inference}. Compare with that, our approach gains effectiveness as is shown in Figure \ref{fig:compare_base}.  Intuitively, adopting an individualized payment rule with a continuous action space allows our mechanism to optimize much better towards the utility objective. 
Note that our algorithm outperforms the state-of-the-art method (Gibbs sampling with DQN) even without an oracle and the importance sampling.

\begin{figure}[th]
    \centering
    \begin{subfigure}[b]{0.22\textwidth}
         \centering
         \includegraphics[width=\textwidth]{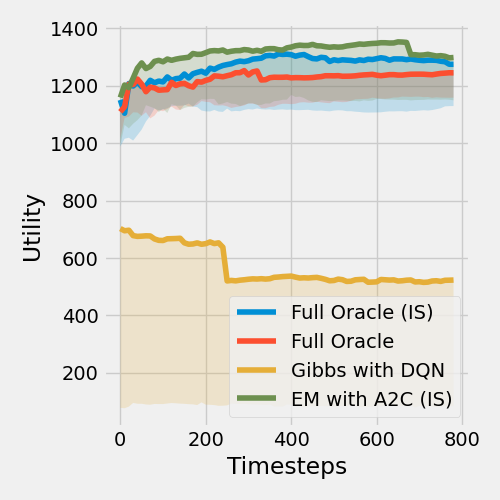}
         \caption{Comparison to Gibbs sampling with DQN \cite{hu2018inference} on Bluebirds with 50 tasks per timestep.\newline}
         \label{fig:compare_2}
     \end{subfigure}
     \hfill
     \begin{subfigure}[b]{0.22\textwidth}
         \centering
         \includegraphics[width=\textwidth]{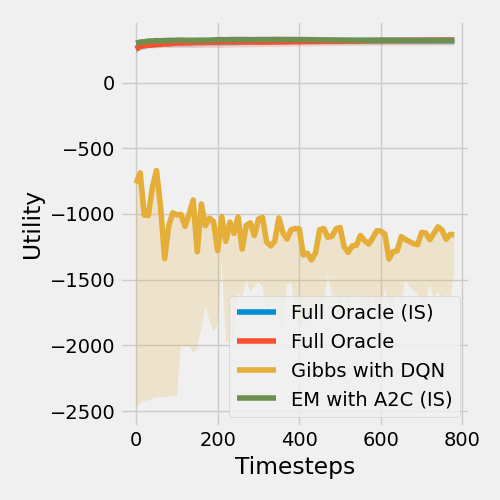}
         \caption{Comparison to Gibbs sampling with DQN \cite{hu2018inference} on Crowdsourced Amazon Sentiment with 100 tasks per timestep.}
         \label{fig:compare_3}
     \end{subfigure}
     \caption{}
\end{figure}

With more tasks per timestep, such as 50 tasks per timestep with the Blubirds dataset, our method significantly outperforms the previous approach, Gibbs Sampling with DQN \cite{hu2018inference}, as shown in Figure \ref{fig:compare_2}. Moreover, the utility achieved by our methods (EM with A2C) is roughly five times (almost 1400) the utility it achieved with 10 tasks per timestep (around 270) showed in Figure \ref{fig:compare_base}. In comparison, Gibbs sampling with DQN achieved only around 200 with 10 tasks per timestep and around 500-600 with 50 tasks per time step. This attests that our method is robust with various environments. 

On larger datasets such as Crowdsourced Amazon Sentiment with $7,803$ answers from $284$ workers, our method quickly learns an incentivizing payment strategy while our baseline, Gibbs sampling with DQN, fails to achieve a positive averaged utility.

Beyond this significant improvement in performance achieved by our approach in a benign environment, our approach is also shown to be effective in much complicated, potentially malicious environments, including ones with collusion. 
On larger datasets such as Crowdsourced Amazon Sentiment with $7,803$ answers, Figure \ref{fig:amazon_is} shows the effectiveness of our proposed approach (partial oracle) under collusion against baselines that are without an oracle or without IS. This result demonstrates the robustness of our approach gained through the availability of oracle under variants of worker strategies, regardless of the size of the datasets.

With the Sentiment Popularity - AMT dataset, our approach with partial oracle is not only shown to be  robust under collusion, but also achieves similar performance as if the oracle is queried in every timestep (Figure \ref{fig:binaryAMT}). When such availability is infeasible as is in most practical applications, our approach remains as effective, which allows the algorithms to be deployed onto a wider range of real-world crowdsourcing systems.

Lastly, our approach is flexible with multi-label datasets such as Weather Sentiment - AMT. Figure \ref{fig:amt_multi} shows that on this multi-label dataset, both importance sampled experience replay and partial oracle are essential for the performance under collusion. Moreover, without the constant availability of an oracle and consequently less burden on the payment, our approach is shown to be most effective when oracle is made available as an action.

\subsection{Bridging theoretical and empirical metrics}

In game theory and peer prediction literature, individual rationality (IR) and incentive compatibility (IC) are two important metrics to assess the mechanism. On a high level, a mechanism that achieves IR and IC should ensure that all participating workers are strictly better off when an honest response is submitted and have no incentive of dropping out of the framework. Notions extending from IC and IR, like strategyproof and group strategyproof, are used to describe the mechanism's capability to overcome potential collusion activities. Here we give formal definitions of IC, IR, strategyproof, and group strategyproof in the context of crowdsourcing and show how these notions are empirically satisfied in our experiments.
\begin{defn}[Individual rationality]
    For all worker $i \in [N]$, the utility gained through participating in crowdsourcing labeling is non-negative. 
\end{defn}
In the context of crowdsourcing, let workers' response be the generated labels. For response $r$ and $r'$, $r' < r$ means label $r$ is more accurate than $r'$. Then incentive compatibility is defined as follows.
\begin{defn}[Incentive compatibility]
    For any worker $i \in [N]$ with response $r' < r$, the utility gained through submitting $r'$ is strictly less than the utility gain through submitting $r$. 
\end{defn}
In other words, a desired mechanism should ensure that honest workers receive better utility in expectation. Moreover, a desired mechanism should ensure that it is in the workers' interests to participate in crowdsourcing. A stronger notion, strategyproof, is introduced to incentivize a worker to not participate in a collusion activity when it takes place.

\begin{defn}[Strategyproof]
    Let $\mathbb{E}[U(e)]$ denotes the expected utility when effort level $e$ is exerted.  A mechanism is strategyproof if for any worker $i$ with the ability to exert effort level $e$, $e'$, $e < e'$, $\mathbb{E} [U(e)] < \mathbb{E}[U(e')]$.
\end{defn}

The above definition only helps to describe a mechanism that is resistant to individual non-rational workers. To define a mechanism that is resistant to group collusion, we need the following definition of group strategyproof.

\begin{defn}[Group strategyproof]
    Let $\mathbb{E} [U(e)]$ denote the expected utility when effort level $e$ is exerted and $N$ be the population of workers.  A mechanism is said to be strategyproof if for any worker subgroup $G \subseteq [N]$ with the ability to exert effort level $e$, $e'$, $e < e'$, we have $\sum_{i \in G}\mathbb{E} [U(e)] < \sum_{i \in G} \mathbb{E}_T [U(e')]$ for any $t \in [T]$.
\end{defn}

Note that for strategyproof and group strategyproof, the definition is with respect to an adaptive adversarial strategy. With our sequential formulation of the problem and our learning approach, it is hard to ensure per step IR, IC, strategyproof, and group strategyproof. However, all of these metrics can be reflected through the fluctuation of the platform's utility over the time horizon. Thus, we discuss the above-mentioned metrics in the long-term expected form over a finite time horizon. 

\paragraph{Long term expected IR and IC} When the long term expected utility of workers becomes negative (thus not IR) or the workers are rewarded better for fewer quality labels in expectation, it is reasonable to deduce that rational workers in the crowd are given no incentive to submit quality labels. According to our general definition of a platform's utility, the platform should also receive a negative expected utility or have no significant improvement of utility over the time horizon (as the quality of labels now only depends on irrational workers). A negative utility is not observed in our approach but is observed in some baseline methods.

\paragraph{Long term expected strategyproof and group strategyproof}
Intuitively, any successful adversarial individual or group strategy would cause a significant decrease in the utility of the platform. Moreover, collusion may indeed be a deceptive enough move and should cause great harm to the platform utility without the existence of an external oracle.

\paragraph{Bayesian Nash equilibrium}
Previous related works such as \cite{liu2016learning} characterize the Bayesian Nash equilibrium (BNE) and show that at equilibrium there is a unique strategy for positive effort exertion. With our deep reinforcement learning approach, though we cannot prove the uniqueness of our strategy at equilibrium, our results can be inferred that reinforcement learning does learn an incentive mechanism that approximately achieves Nash equilibrium. This is reflected through our increasing flat curve towards the end of the training, where the utility of the platform encounters very little changes as a result of little change in inferred label accuracy from workers.

\section{Conclusion and future work}
We propose a reinforcement learning based approach for the problem of sequential eliciting information without verification (EIWV) for an heterogeneous worker group of unknown cost distribution, potential irrationality and collusion. Our method is shown to be effective in a wide range of settings and robust to collusion with the aid of a costly oracle. We provide comprehensive experiments to validate our effectiveness on a variety of datasets. Our approach outperforms the baseline methods, with an especially large margin on large datasets.

An interesting direction of future work is to consider the fairness of the payment mechanism, which has a significant effect on the long-term prosperity of the crowdsourcing platform. Considering that the workers share some information among them, their willingness of exerting efforts and being truthful will be affected by comparing their utility with others. A fair distribution of payment will be the key to maintain the workers' productivity in a long run.

\bibliographystyle{aaai}
\bibliography{ref.bib}

\begin{thebibliography}{}

\bibitem[\protect\citeauthoryear{Blitzer, Dredze, and
  Pereira}{2007}]{blitzer2007biographies}
Blitzer, J.; Dredze, M.; and Pereira, F.
\newblock 2007.
\newblock Biographies, bollywood, boom-boxes and blenders: Domain adaptation
  for sentiment classification.
\newblock In {\em Proceedings of the 45th Annual Meeting of the Association of
  computational linguistics},  440--447.

\bibitem[\protect\citeauthoryear{Checco, Bates, and
  Demartini}{2019}]{checco2019quality}
Checco, A.; Bates, J.; and Demartini, G.
\newblock 2019.
\newblock Quality control attack schemes in crowdsourcing.
\newblock In {\em IJCAI International Joint Conference on Artificial
  Intelligence}, volume 2019,  6136--6140.
\newblock AAAI Press.

\bibitem[\protect\citeauthoryear{Cruz~Jr}{1975}]{cruz1975survey}
Cruz~Jr, J.
\newblock 1975.
\newblock Survey of nash and stackelberg equilibrim strategies in dynamic
  games.
\newblock In {\em Annals of Economic and Social Measurement, Volume 4, number
  2}. NBER.
\newblock  339--344.

\bibitem[\protect\citeauthoryear{Dasgupta and
  Ghosh}{2013}]{dasgupta2013crowdsourced}
Dasgupta, A., and Ghosh, A.
\newblock 2013.
\newblock Crowdsourced judgement elicitation with endogenous proficiency.
\newblock In {\em Proceedings of the 22nd International Conference on World
  Wide Web},  319--330.

\bibitem[\protect\citeauthoryear{Difallah \bgroup et al\mbox.\egroup
  }{2015}]{difallah2015dynamics}
Difallah, D.~E.; Catasta, M.; Demartini, G.; Ipeirotis, P.~G.; and
  Cudr{\'e}-Mauroux, P.
\newblock 2015.
\newblock The dynamics of micro-task crowdsourcing: The case of amazon mturk.
\newblock In {\em Proceedings of the 24th International Conference on World
  Wide Web},  238--247.

\bibitem[\protect\citeauthoryear{Haarnoja \bgroup et al\mbox.\egroup
  }{2018}]{haarnoja2018soft}
Haarnoja, T.; Zhou, A.; Abbeel, P.; and Levine, S.
\newblock 2018.
\newblock Soft actor-critic: Off-policy maximum entropy deep reinforcement
  learning with a stochastic actor.
\newblock In {\em International Conference on Machine Learning},  1861--1870.
\newblock PMLR.

\bibitem[\protect\citeauthoryear{Han \bgroup et al\mbox.\egroup
  }{2021}]{han2021truthful}
Han, Q.; Ruan, S.; Kong, Y.; Liu, A.; Mohsin, F.; and Xia, L.
\newblock 2021.
\newblock Truthful information elicitation from hybrid crowds.
\newblock {\em arXiv preprint arXiv:2107.10119}.

\bibitem[\protect\citeauthoryear{Ho and Vaughan}{2012}]{ho2012online}
Ho, C.-J., and Vaughan, J.
\newblock 2012.
\newblock Online task assignment in crowdsourcing markets.
\newblock In {\em Proceedings of the AAAI Conference on Artificial
  Intelligence}, volume~26.

\bibitem[\protect\citeauthoryear{Ho, Jabbari, and
  Vaughan}{2013}]{pmlr-v28-ho13}
Ho, C.-J.; Jabbari, S.; and Vaughan, J.~W.
\newblock 2013.
\newblock Adaptive task assignment for crowdsourced classification.
\newblock In Dasgupta, S., and McAllester, D., eds., {\em Proceedings of the
  30th International Conference on Machine Learning}, volume~28 of {\em
  Proceedings of Machine Learning Research},  534--542.
\newblock Atlanta, Georgia, USA: PMLR.

\bibitem[\protect\citeauthoryear{Howe and others}{2006}]{howe2006rise}
Howe, J., et~al.
\newblock 2006.
\newblock The rise of crowdsourcing.
\newblock {\em Wired Magazine} 14(6):1--4.

\bibitem[\protect\citeauthoryear{Hu \bgroup et al\mbox.\egroup
  }{2018}]{hu2018inference}
Hu, Z.; Liang, Y.; Zhang, J.; Li, Z.; and Liu, Y.
\newblock 2018.
\newblock Inference aided reinforcement learning for incentive mechanism design
  in crowdsourcing.
\newblock {\em Advances in Neural Information Processing Systems}
  31:5507--5517.

\bibitem[\protect\citeauthoryear{Konda and Tsitsiklis}{2000}]{konda2000actor}
Konda, V.~R., and Tsitsiklis, J.~N.
\newblock 2000.
\newblock Actor-critic algorithms.
\newblock In {\em Advances in Neural Information Processing Systems},
  1008--1014.

\bibitem[\protect\citeauthoryear{Kong and Schoenebeck}{2019}]{10.1145/3296670}
Kong, Y., and Schoenebeck, G.
\newblock 2019.
\newblock An information theoretic framework for designing information
  elicitation mechanisms that reward truth-telling.
\newblock {\em ACM Transaction Economics Computation} 7(1).

\bibitem[\protect\citeauthoryear{Liu and Chen}{2016}]{liu2016learning}
Liu, Y., and Chen, Y.
\newblock 2016.
\newblock Learning to incentivize: Eliciting effort via output agreement.
\newblock In {\em Proceedings of the International Joint Conference on
  Artificial Intelligence}.
\newblock International Joint Conferences on Artificial Intelligence.

\bibitem[\protect\citeauthoryear{Liu and Chen}{2017}]{liu2017sequential}
Liu, Y., and Chen, Y.
\newblock 2017.
\newblock Sequential peer prediction: Learning to elicit effort using posted
  prices.
\newblock In {\em Thirty-First AAAI Conference on Artificial Intelligence}.

\bibitem[\protect\citeauthoryear{Mnih \bgroup et al\mbox.\egroup
  }{2016}]{mnih2016asynchronous}
Mnih, V.; Badia, A.~P.; Mirza, M.; Graves, A.; Lillicrap, T.; Harley, T.;
  Silver, D.; and Kavukcuoglu, K.
\newblock 2016.
\newblock Asynchronous methods for deep reinforcement learning.
\newblock In {\em International Conference on Machine Learning},  1928--1937.
\newblock PMLR.

\bibitem[\protect\citeauthoryear{Oleson \bgroup et al\mbox.\egroup
  }{2011}]{oleson2011programmatic}
Oleson, D.; Sorokin, A.; Laughlin, G.; Hester, V.; Le, J.; and Biewald, L.
\newblock 2011.
\newblock Programmatic gold: Targeted and scalable quality assurance in
  crowdsourcing.
\newblock In {\em Workshops at the Twenty-Fifth AAAI Conference on Artificial
  Intelligence}.

\bibitem[\protect\citeauthoryear{Schaul \bgroup et al\mbox.\egroup
  }{2016}]{schaul2015prioritized}
Schaul, T.; Quan, J.; Antonoglou, I.; and Silver, D.
\newblock 2016.
\newblock Prioritized experience replay.
\newblock In {\em International Conference on Learning Representations}.

\bibitem[\protect\citeauthoryear{Schoenebeck, Yu, and
  Zhang}{2021}]{10.1145/3442381.3449840}
Schoenebeck, G.; Yu, F.-Y.; and Zhang, Y.
\newblock 2021.
\newblock Information elicitation from rowdy crowds.
\newblock In {\em Proceedings of the Web Conference},  3974–3986.
\newblock New York, NY, USA: Association for Computing Machinery.

\bibitem[\protect\citeauthoryear{Shah and Zhou}{2015}]{shah2015double}
Shah, N.~B., and Zhou, D.
\newblock 2015.
\newblock Double or nothing: Multiplicative incentive mechanisms for
  crowdsourcing.
\newblock {\em Advances in Neural Information Processing Systems} 28:1--9.

\bibitem[\protect\citeauthoryear{Shah and Zhou}{2016}]{pmlr-v48-shaha16}
Shah, N., and Zhou, D.
\newblock 2016.
\newblock No oops, you won't do it again: Mechanisms for self-correction in
  crowdsourcing.
\newblock In Balcan, M.~F., and Weinberger, K.~Q., eds., {\em Proceedings of
  The 33rd International Conference on Machine Learning}, volume~48 of {\em
  Proceedings of Machine Learning Research},  1--10.
\newblock New York, New York, USA: PMLR.

\bibitem[\protect\citeauthoryear{Shah and Zhou}{2020}]{10.1145/3396863}
Shah, N.~B., and Zhou, D.
\newblock 2020.
\newblock Approval voting and incentives in crowdsourcing.
\newblock {\em ACM Transaction on Economics Computation} 8(3).

\bibitem[\protect\citeauthoryear{Shah, Zhou, and
  Peres}{2015}]{pmlr-v37-shaha15}
Shah, N.; Zhou, D.; and Peres, Y.
\newblock 2015.
\newblock Approval voting and incentives in crowdsourcing.
\newblock In {\em Proceedings of the 32nd International Conference on Machine
  Learning},  10--19.
\newblock Lille, France: PMLR.

\bibitem[\protect\citeauthoryear{Shnayder \bgroup et al\mbox.\egroup
  }{2016a}]{10.1145/2940716.2940790}
Shnayder, V.; Agarwal, A.; Frongillo, R.; and Parkes, D.~C.
\newblock 2016a.
\newblock Informed truthfulness in multi-task peer prediction.
\newblock In {\em Proceedings of the 2016 ACM Conference on Economics and
  Computation}, EC '16,  179–196.
\newblock New York, NY, USA: Association for Computing Machinery.

\bibitem[\protect\citeauthoryear{Shnayder \bgroup et al\mbox.\egroup
  }{2016b}]{shnayder2016informed}
Shnayder, V.; Agarwal, A.; Frongillo, R.; and Parkes, D.~C.
\newblock 2016b.
\newblock Informed truthfulness in multi-task peer prediction.
\newblock In {\em Proceedings of the 2016 ACM Conference on Economics and
  Computation},  179--196.

\bibitem[\protect\citeauthoryear{Simpson \bgroup et al\mbox.\egroup
  }{2015}]{simpson2015language}
Simpson, E.~D.; Venanzi, M.; Reece, S.; Kohli, P.; Guiver, J.; Roberts, S.~J.;
  and Jennings, N.~R.
\newblock 2015.
\newblock Language understanding in the wild: Combining crowdsourcing and
  machine learning.
\newblock In {\em Proceedings of the 24th International Conference on World
  Wide Web},  992--1002.

\bibitem[\protect\citeauthoryear{Venanzi \bgroup et al\mbox.\egroup
  }{2015a}]{soton376544}
Venanzi, M.; Teacy, W.; Rogers, A.; and Jennings, N.
\newblock 2015a.
\newblock Sentiment popularity - amazon mechanical turk dataset.
\newblock Supports: Venanzi, Matteo, Teacy, W.T.L., Rogers, Alex and Jennings,
  Nicholas R. (2015) Bayesian modelling of community-based multidimensional
  trust in participatory sensing under data sparsity. In, International Joint
  Conference on Artificial Intelligence (IJCAI-15), Buenos Aires, AR, 25 - 31
  Jul 2015. 8pp, 717-724.

\bibitem[\protect\citeauthoryear{Venanzi \bgroup et al\mbox.\egroup
  }{2015b}]{soton376543}
Venanzi, M.; Teacy, W.; Rogers, A.; and Jennings, N.
\newblock 2015b.
\newblock Weather sentiment - amazon mechanical turk dataset.

\bibitem[\protect\citeauthoryear{Welinder \bgroup et al\mbox.\egroup
  }{2010}]{WelinderEtal10b}
Welinder, P.; Branson, S.; Belongie, S.; and Perona, P.
\newblock 2010.
\newblock {The Multidimensional Wisdom of Crowds}.
\newblock In {\em Advances in Neural Information Processing Systems}.

\bibitem[\protect\citeauthoryear{Witkowski and
  Parkes}{2012}]{10.1145/2229012.2229085}
Witkowski, J., and Parkes, D.~C.
\newblock 2012.
\newblock Peer prediction without a common prior.
\newblock In {\em Proceedings of the 13th ACM Conference on Electronic
  Commerce}, EC '12,  964–981.
\newblock New York, NY, USA: Association for Computing Machinery.

\bibitem[\protect\citeauthoryear{Witkowski and
  Parkes}{2013}]{witkowski2013learning}
Witkowski, J., and Parkes, D.~C.
\newblock 2013.
\newblock Learning the prior in minimal peer prediction.
\newblock In {\em Proceedings of the 3rd Workshop on Social Computing and User
  Generated Content at the ACM Conference on Electronic Commerce}, volume~14.

\bibitem[\protect\citeauthoryear{Witkowski \bgroup et al\mbox.\egroup
  }{2013}]{witkowski2013dwelling}
Witkowski, J.; Bachrach, Y.; Key, P.; and Parkes, D.
\newblock 2013.
\newblock Dwelling on the negative: Incentivizing effort in peer prediction.
\newblock In {\em Proceedings of the AAAI Conference on Human Computation and
  Crowdsourcing}, volume~1.

\bibitem[\protect\citeauthoryear{Yang, Cai, and Zheng}{2018}]{8693123}
Yang, P.; Cai, H.; and Zheng, Z.
\newblock 2018.
\newblock Improving the quality of crowdsourcing labels by combination of
  golden data and incentive.
\newblock In {\em 2018 12th IEEE International Conference on
  Anti-counterfeiting, Security, and Identification (ASID)},  10--15.

\bibitem[\protect\citeauthoryear{Zhao \bgroup et al\mbox.\egroup
  }{2020}]{zhao2020preference}
Zhao, Y.; Zheng, K.; Yin, H.; Liu, G.; Fang, J.; and Zhou, X.
\newblock 2020.
\newblock Preference-aware task assignment in spatial crowdsourcing: from
  individuals to groups.
\newblock {\em IEEE Transactions on Knowledge and Data Engineering}.

\bibitem[\protect\citeauthoryear{Zheng \bgroup et al\mbox.\egroup
  }{2017}]{10.14778/3055540.3055547}
Zheng, Y.; Li, G.; Li, Y.; Shan, C.; and Cheng, R.
\newblock 2017.
\newblock Truth inference in crowdsourcing: Is the problem solved?
\newblock {\em Proceedings of the VLDB Endowment} 10(5):541–552.

\end{thebibliography}

\clearpage
\onecolumn
\appendix

\section{Datasets}

\paragraph{Bluebirds \cite{WelinderEtal10b}} 
This dataset contains both individual answers from 108 workers and the ground truth labels for 38 tasks regarding whether there is a blue bird in the picture. The dataset only contains tasks with binary labels and includes $4,212$ total answers from workers. 

\paragraph{Crowdsourced Amazon Sentiment \cite{blitzer2007biographies}}
The Crowdsourced Amazon Sentiment dataset includes $7,803$ answers from 284 workers for $1,011$ Amazon reviews to decide whether the review is about a book or not. The tasks are binary-labeled tasks and have ground truth labels.

\paragraph{Sentiment Popularity - AMT \cite{soton376544}}
This is 
the largest binary labeled dataset we studied, with $10,000$ answers from 143 workers and 500 tasks on deciding whether a movie review is positive or negative. Each task has a ground truth label provided by the review website. The crowdsourced labels are from Amazon Mechanical Turk. 

\paragraph{Weather Sentiment - AMT \cite{soton376543}} 
This multi-labeled tasks dataset contains $6,000$ answers from 110 workers on 330 tasks on classifying weathers into five categories by looking into a tweet: negative (0), neutral (1), positive (2), tweet not related to weather (3) and can't tell (4). The labeled answers are data from Amazon Mechanical Turk and every task has a ground truth label. 

\section{Flexibility of our framework}
\begin{figure*}[!htbp]\label{test}
\begin{minipage}[t]{0.25\textwidth}
  \includegraphics[width=\linewidth]{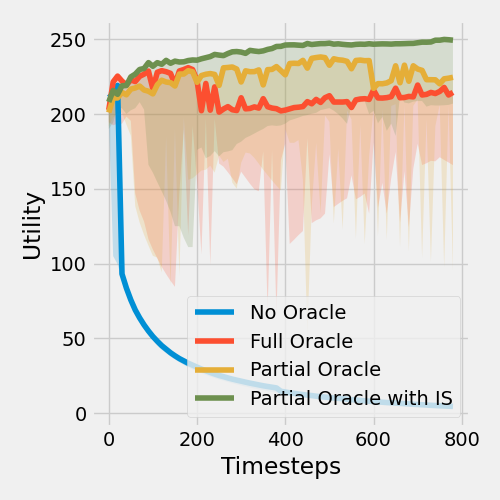}
  \subcaption{}
  \label{fig:utility}
\end{minipage}%
\hfill 
\begin{minipage}[t]{0.25\textwidth}
  \includegraphics[width=\linewidth]{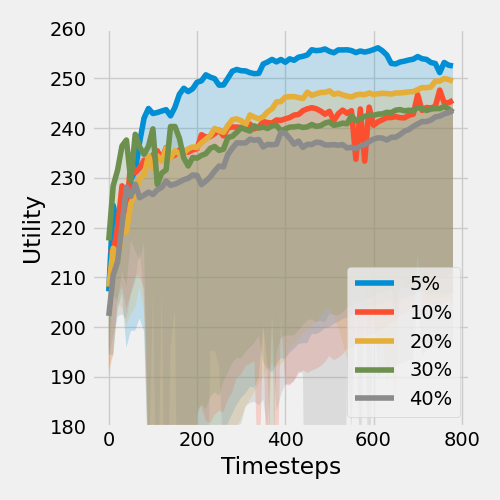}
  \subcaption{}
  \label{fig:diff_cost}
\end{minipage}%
\hfill
\begin{minipage}[t]{0.25\textwidth}
  \includegraphics[width=\linewidth]{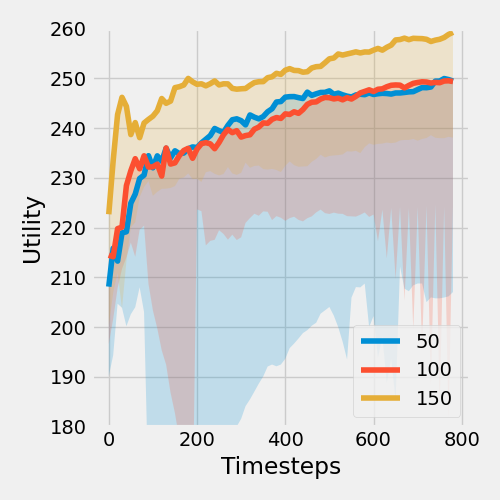}
  \subcaption{}
  \label{fig:diff_coll_rate}
\end{minipage}%
\hfill
\begin{minipage}[t]{0.25\textwidth}
    \centering
  \includegraphics[width=\linewidth]{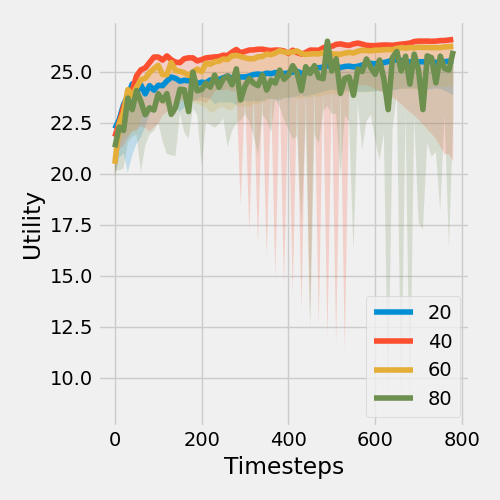}
  \subcaption{}
  \label{fig:diff_task}
\end{minipage}%
\hfill

\caption{Figure \ref{fig:diff_cost} compares the performance with different levels of oracle cost from 5-40 percent more of the chosen payment. Figure \ref{fig:diff_coll_rate} compares the performance with collusion rate of 50, 10, 150 (e.g. start colluding once every 50 timesteps). Figure \ref{fig:diff_task} compares the performance with a different number of tasks per worker at each timestep, ranging from 20 to 80. }
\end{figure*}
For the experimental setting, all experiments for the ablation study are conducted over 800 timesteps on the Bluebirds dataset. Figure \ref{fig:utility} gives an overview of the performance of A2C with/without oracle and with/without importance sampled buffer with 20 percent oracle cost, collusion rate of 50, a maximum payment of 11, maximum effort cost of 10, and 10 tasks per time step. Unless otherwise indicated, all environment configuration remain the same in the following figures. 

Before moving on to ablation results, we first perform a comparison between the effect of each components in our approach on the Bluebirds dataset (Figure \ref{fig:utility}). This will showoff the flexibility of our environment under a variety of environment parameter combinations. The partial query of oracle and importance sampled experience replay are shown to be essential for better performance and stable training under complicated environment with possible collusion.

Figure \ref{fig:diff_cost} shows the effect of the $\alpha$-oracle cost on the performance of the algorithm. Regardless of the increased cost of the oracle, our approach's performance decreases only in a moderate rate. The difference in cost of oracle is a realization of the availability of golden tasks or expert advice, thus proves the feasibility of our approach in a wide range of environments.  

Figure \ref{fig:diff_coll_rate} studies the effect of collusion rate to the performance. By collusion rate of $m$, the worker will conduct a collusion every $m$ timesteps and can only be stopped by giving every participant the lowest possible payment. The performance of an approach is likely to decrease when $m$ decreases. From the figure, A2C with important sampling performance can handle a collusion rate of 150 or larger. The algorithm will have limited performance for lower collusion rates as the reinforcement learning algorithm has limited horizon between collusion points to balance the incentivizing payment strategy and the anti-collusion strategy.

Figure \ref{fig:diff_task} shows the effect of a different number of tasks distributed at each time step. The utility is normalized to per task for a fair comparison. We observe that the number of tasks distributed at each time step does not have a significant impact on the performance. 

\end{document}